\documentclass[11pt,a4paper]{article}
\usepackage[hyperref]{emnlp2020}
\pdfoutput=1
\usepackage{times}
\usepackage{latexsym}

\usepackage{times}
\usepackage{latexsym}
\usepackage{amsmath}
\usepackage{url}
\usepackage{amssymb}
\usepackage{amsfonts}
\usepackage{graphicx}
\usepackage{tabularx}
\usepackage{multirow}
\usepackage{arydshln}
\usepackage{mathtools,nccmath}

\usepackage{tcolorbox}

\usepackage{enumitem}
\aclfinalcopy % Uncomment this line for the final submission
\title{WNUT-2020 Task 2: \\ Identification of Informative COVID-19 English Tweets}

\author{Dat Quoc Nguyen$^{1,\ast}$, Thanh Vu$^{2,}$\thanks{\ \ The first two authors contributed equally to this work. Most of the work was done when Thanh Vu was at the Australian e-Health Research Centre, CSIRO, Australia.}\ , Afshin Rahimi$^3$, Mai Hoang Dao$^1$, \\ \textbf{Linh The Nguyen}$^1$ \and \textbf{Long Doan}$^1$ \\
  $^1$VinAI Research, Vietnam; $^2$Oracle Digital Assistant, Oracle, Australia; \\ $^3$The University of Queensland, Australia\\
   {\normalsize \tt{v.datnq9@vinai.io; thanh.v.vu@oracle.com; a.rahimi@uq.edu.au}}\\
   {\normalsize \tt{\{v.maidh3, v.linhnt140, v.longdct\}@vinai.io}}
   }

\date{}

\begin{document}
\maketitle
\begin{abstract}
In this paper, we provide an overview of the WNUT-2020 shared task on the identification of informative COVID-19 English Tweets. We describe how we construct a corpus of 10K Tweets and organize the development and evaluation phases for this task. In addition, we also present a brief summary of results obtained from the final system evaluation submissions of 55 teams, finding that (i) many systems obtain very high performance, up to 0.91 F$_1$ score, (ii) the majority of the submissions achieve substantially higher results than the baseline fastText \cite{joulin2017bag}, and (iii) fine-tuning pre-trained language models on relevant language data followed by supervised training performs well in this task.
\end{abstract}

\section{Introduction}\label{sec:intro}

As of late-September 2020, the COVID-19 Coronavirus pandemic has led to about 1M deaths and 33M infected patients from  213 countries and territories, creating fear and panic for people all around the world.\footnote{\url{https://www.worldometers.info/coronavirus/}} Recently, much attention has been paid to building monitoring systems (e.g. The Johns Hopkins Coronavirus Dashboard) to track the development of the pandemic and to provide users the information related to the virus,\footnote{\url{https://coronavirus.jhu.edu/map.html}} e.g. any new suspicious/confirmed cases near/in the users' regions. 

It is worth noting  that most of the ``official'' sources used in the tracking tools are not frequently kept up to date with the current pandemic situation, e.g. WHO updates the pandemic information only once a day. Those monitoring systems thus use social network data, e.g. from Twitter, as a real-time alternative source for updating the pandemic information, generally by crowd-sourcing or searching for related information manually. However, the pandemic has been spreading rapidly; we observe a massive amount of data on social networks, e.g. about 3.5M of COVID-19 English Tweets posted daily on the Twitter platform \cite{datacovid} in which the majority are uninformative. %\footnote{\url{https://ieee-dataport.org/open-access/coronavirus-covid-19-tweets-dataset}}  
Thus, it is important to be able to select the informative Tweets (e.g.\ COVID-19 Tweets related to new cases or suspicious cases) for downstream applications. However, manual approaches to identify the informative Tweets require significant human efforts, do not scale with rapid developments, and are costly.

To help handle the problem, \textit{we propose a shared task which is to automatically identify whether a COVID-19 English Tweet is informative or not}.   Our task is defined as a binary classification problem: Given an English Tweet related to COVID-19, decide whether it should be classified as  INFORMATIVE or UNINFORMATIVE. Here, informative Tweets provide information about  suspected, confirmed, recovered and death cases as well as the location or travel history of the cases. The following example presents an informative Tweet:

\medskip

\begin{tcolorbox}[title=INFORMATIVE]
Update: Uganda Health Minister Jane Ruth Aceng has confirmed the first \#coronavirus case in Uganda. The patient is a 36-year-old Ugandan male who arrived from Dubai today aboard Ethiopian Airlines. Patient travelled to Dubai 4 days ago. \#CoronavirusPandemic
\end{tcolorbox}

\medskip

The goals of our shared task are: (i) To develop a language processing task that potentially impacts research and downstream applications, and (ii) To provide the research community with a new dataset for identifying informative COVID-19 English Tweets. 
To achieve the goals, we manually construct a dataset of 10K COVID-19 English Tweets with INFORMATIVE and UNINFORMATIVE labels. We   believe that  the  dataset and systems developed for our task will be beneficial for the development of COVID-19 monitoring systems.  
All practical information, data download links and the final evaluation results can be found at  the CodaLab website of our shared task: \url{https://competitions.codalab.org/competitions/25845}.

\section{The WNUT-2020 Task 2 dataset}

\subsection{Annotation guideline}
We define the guideline to annotate a COVID-19 related Tweet with the ``INFORMATIVE'' label if the Tweet mentions suspected cases, confirmed cases, recovered cases, deaths, number of tests performed as well as location or travel history associated with the confirmed/suspected cases. %We believe such information is crucial for the development of COVID-19 related monitoring systems. 
%Another example of informative Tweet is as follows:

%\medskip
%
%\begin{tcolorbox}[title=INFORMATIVE]
%MORE RECOVERIES! The Department of Health reports 7 new COVID-19 recoveries, bringing the total number of recovered patients to 49. Please stay at home! \#COVID19 HTTPURL
%%\bigskip
%
%%Update: Uganda Health Minister Jane Ruth Aceng has confirmed the first \#coronavirus case in Uganda. The patient is a 36-year-old Ugandan male who arrived from Dubai today aboard Ethiopian Airlines. Patient travelled to Dubai 4 days ago. \#CoronavirusPandemic
%\end{tcolorbox}
%
%\medskip

In addition, we also set further requirements in which the ``INFORMATIVE'' Tweet has to satisfy. In particular, the ``INFORMATIVE'' Tweet should not present a rumor or prediction. Furthermore, quantities mentioned in the Tweet have to be specific (e.g. ``\emph{two new cases}'' or ``\emph{about 125 tested positives}'') or could be inferred directly (e.g. ``\emph{120 coronavirus tests done so far, 40\% tested positive}''), but not purely in percentages or rates (e.g. ``\emph{20\%}'', ``\emph{1000 per million}'', or ``\emph{a third}''). 

The COVID-19 related Tweets not satisfying the ``INFORMATIVE'' annotation guideline are annotated with the ``UNINFORMATIVE'' label. An uninformative Tweet example is as follows:

\medskip
\begin{tcolorbox}[title=UNINFORMATIVE]
Indonesia frees 18,000 inmates, as it records highest \#coronavirus death toll in Asia behind China HTTPURL
\end{tcolorbox}
%\medskip

\subsection{COVID-19 related Tweet collection}

To be able to construct a dataset used in our shared task, we first have to crawl the COVID-19 related Tweets.  We collect a general Tweet corpus related to the COVID-19 pandemic based on a predefined list of 10 keywords, including: ``coronavirus'', ``covid-19'', ``covid\_19'', ``covid\_2019'', ``covid19'', ``covid2019'', ``covid-2019'', ``CoronaVirusUpdate'', ``Coronavid19'' and ``SARS-CoV-2''. We utilize the Twitter streaming API to download real-time English Tweets containing at least one keyword from the predefined list.\footnote{\url{https://developer.twitter.com/en/docs/twitter-api/v1/tweets/filter-realtime/overview}} 

We stream the Tweet data for four months using the API from 01$^{st}$ March 2020 to 30$^{th}$ June 2020. 
We then filter out Tweets containing less than 10 words (including hashtags and user mentions) as well as Tweets from users with less than five hundred followers. This is to help reduce the rate of Tweets with fake news (our manual annotation process does not involve in verifying fake news) with a rather strong assumption that reliable information is more likely to be propagated by users with a large number of followers.\footnote{We acknowledge that there are accounts with a large number of followers, who participate in publication and propagation of misinformation.} To handle the duplication problem: (i) we remove Retweets starting with the ``RT'' token, and (ii) in cases where two Tweets are the same after lowecasing as well as removing hashtags and user mentions, the earlier Tweet is kept and the subsequent Tweet will be filtered out as it tends to be a Retweet. Applying these filtering steps results in a final corpus of about 23M COVID-19 English Tweets. % Here, user mentions and web/url links are converted into special normalized forms ``@USER'' and ``HTTPURL'', respectively

\subsection{Annotation process}

%\subsubsection*{Phase 1}
From the corpus of 23M Tweets, we select Tweets which are potentially informative, containing predefined strings relevant to the annotation guideline such as ``confirm'', ``positive'', ``suspected'', ``death'', ``discharge'', ``test'' and ``travel history''. We then remove similar Tweets with the token-based cosine similarity score \cite{5767865} that is  equal or greater than 0.7, resulting in a dataset of ``INFORMATIVE'' candidates. We then  randomly sample 2K Tweets from this dataset for the first phase of annotation. 

Three annotators are employed to independently annotate each of the 2K Tweets with one of the two labels ``INFORMATIVE'' and ``UNINFORMATIVE''. We use the ``docanno'' toolkit  for handling the annotations \cite{doccano}. We measure the inter-annotator agreement to assess the quality of annotations and to see whether the guideline allows to carry out the task consistently. In particular, we use the Fleiss' Kappa coefficient to assess the annotator agreement \cite{fleiss1971measuring}. For this first phase, the Kappa score is 0.797 which can be interpreted as {substantial} \cite{landis1977measurement}. We further run a discussion for Tweets where there is a disagreement in the assigned labels among the annotators. The discussion is to determine the final labels of the Tweets as well as to improve the quality of the annotation guideline. 

%\subsubsection*{Phase 2}
For the second phase, we employ the 2K annotated Tweets from the first phase to train a binary fastText classifier \cite{joulin2017bag} to classify a COVID-19 related Tweet into either ``INFORMATIVE'' or ``UNINFORMATIVE''. We utilize the trained classifier to predict the probability of ``INFORMATIVE'' for each of  all remaining Tweets in the dataset of ``INFORMATIVE'' candidates from the first phase. Then we randomly sample 8K Tweets from the candidate dataset, including 3K, 2K and 3K Tweets associated with the probability $\in$ [0.0, 0.3), [0.3, 0.7) and [0.7, 1.0], respectively (here, we do not sample from the existing 2K annotated Tweets). The goal here is to select Tweets with varying degree of detection difficulty (with respect to the baseline) in both labels.

The three annotators then independently assign the ``INFORMATIVE'' or ``UNINFORMATIVE'' label  to each of the 8K Tweets. The Kappa score is obtained at 0.818 which can be interpreted as {almost perfect} \cite{landis1977measurement}. Similar to the first phase, for each Tweet with a disagreement among the annotators, we also run a further discussion to decide its final label annotation. 

We merge the two datasets from the first and second phases to formulate the final gold standard corpus of 10K annotated Tweets, consisting of 4,719 ``INFORMATIVE'' Tweets and 5,281 ``UNINFORMATIVE'' Tweets. 

\begin{table}[!t]
\centering
\resizebox{7.5cm}{!}{
\begin{tabular}{l|lll|l}
\hline
Item & Training & Validation & Test & Total \\\hline
\#INFOR & 3,303 & 472 & 944 & 4,719\\
\#UNINF & 3,697 & 528 & 1,056 & 5,281\\\hline
Total & 7,000 & 1,000 & 2,000 & 10,000\\\hline

\end{tabular}
}
\caption{Basic statistics of our dataset. \#INFOR and \#UNINF denote the numbers of ``INFORMATIVE''  and ``UNINFORMATIVE'' Tweets, respectively.}
\label{tab:data}
\end{table}

\subsection{Data partitions}

To split the gold standard corpus into training, validation and test sets, we first categorize its Tweets into two categories of ``easy'' and ``not-easy'', in which the ``not-easy'' category contains Tweets with a label disagreement among annotators before participating in  the annotation discussions. We then randomly select 7K Tweets for training, 1K Tweets for validation and 2K Tweets for test  with a constraint that ensures the number of the ``not-easy'' Tweets in the training is equal to that in the validation and test sets. Table \ref{tab:data} describes the basic statistics of our corpus.

\section{Task organization}

\paragraph{Development phase:}
Both the training and validation sets with gold labels are released publicly  to all participants for system development. Although we provide a default training and validation split of the released data, participants are free to use this data in any way they find useful when training and tuning their systems, e.g. using a different split or performing cross-validation. 

\paragraph{Evaluation  phase:} 
The raw test set is released when the final phase of system evaluation  starts. To keep fairness among participants, the raw test set is a relatively large set of 12K Tweets, and the actual 2K test Tweets by which the participants' system outputs are evaluated are hidden in this large test set. We allow each  participant to upload at most 2 submissions during this  final evaluation phase, in which the submission obtaining higher F$_1$ score is ranked higher in the leaderboard. 

\paragraph{Metrics:} Systems are evaluated using standard evaluation metrics, including Accuracy, Precision, Recall and  F$_1$ score. Note that the latter three metrics of Precision, Recall and  F$_1$ will be calculated for the ``INFORMATIVE'' label only. The system evaluation submissions are ranked by the F$_1$ score.

%\medskip
%\begin{eqnarray*}
%\text{Accuracy} &=& \dfrac{\mid\text{True Positives}\mid+\mid\text{True Negatives}\mid}{\text{Total number of instances}} \\
%\\
%\text{Precision} &=& \dfrac{\mid\text{True Positives}\mid}{\mid\text{True Positives}\mid+\mid\text{False Positives}\mid}\\
%\\
%\text{Recall} &=& \dfrac{\mid\text{True Positives}\mid}{\mid\text{True Positives}\mid+\mid\text{False Negatives}\mid}\\
%\\
%\text{F}_1 &=& 2 \times  \dfrac{\text{Precision} \times \text{Recall}}{\text{Precision}+\text{Recall}}
%\end{eqnarray*}
%\medskip
%\noindent 

\paragraph{Baseline:}  fastText \cite{joulin2017bag} is used as our baseline, employing the default data split.

\begin{table*}[!t]
\centering
\resizebox{16cm}{!}{
\def\arraystretch{1.125}
\setlength{\tabcolsep}{0.3em}
\begin{tabular}{l|llll||l|llll}
\hline
\textbf{Team} & \textbf{F$_1$} & \textbf{P} & \textbf{R} & \textbf{Acc.} &   \textbf{Team} & \textbf{F$_1$} & \textbf{P} & \textbf{R} & \textbf{Acc.}  \\
\hline
NutCracker & \textbf{0.9096} & 0.9135 & 0.9057 & \textbf{0.9150} & CUBoulder-UBC & 0.8841 & 0.8606 & 0.9089 & 0.8875 \\
NLP\_North & \textbf{0.9096} & 0.9029 & 0.9163 & 0.9140 & Sic Mundus & 0.8823 & 0.8832 & 0.8814 & 0.8890 \\
UIT-HSE & 0.9094 & 0.9046 & 0.9142 & 0.9140 & LynyrdSkynyrd & 0.8805 & 0.8567 & 0.9057 & 0.8840 \\
\#GCDH & 0.9091 & 0.8919 & 0.9269 & 0.9125 & Dartmouth CS & 0.8757 & 0.8818 & 0.8697 & 0.8835 \\
Loner & 0.9085 & 0.8918 & 0.9258 & 0.9120 & L3STeam & 0.8754 & 0.8654 & 0.8856 & 0.8810 \\
Phonemer & 0.9037 & 0.8934 & 0.9142 & 0.9080 & XSellResearch & 0.8739 & 0.8857 & 0.8623 & 0.8825 \\
EdinburghNLP & 0.9011 & 0.8768 & 0.9269 & 0.9040 & Linguist Geeks & 0.8715 & 0.9130 & 0.8337 & 0.8840 \\
TATL & 0.9008 & 0.8588 & \textbf{0.9470} & 0.9015 & DSC-IITISM & 0.8715 & 0.8343 & 0.9121 & 0.8730 \\
SunBear & 0.9005 & 0.8728 & 0.9301 & 0.9030 & AmazingAI & 0.8714 & 0.8637 & 0.8792 & 0.8775 \\
InfoMiner & 0.9004 & 0.9102 & 0.8909 & 0.9070 & Siva & 0.8527 & 0.8115 & 0.8983 & 0.8535 \\
NEU & 0.8992 & 0.8959 & 0.9025 & 0.9045 & CSECU-DSG & 0.8198 & 0.8155 & 0.8242 & 0.8290 \\
Not-NUTs & 0.8991 & 0.8787 & 0.9206 & 0.9025 & IIITBH & 0.7979 & 0.7991 & 0.7966 & 0.8095 \\
UET & 0.8989 & 0.8891 & 0.9089 & 0.9035 & NLPRL & 0.7854 & 0.8335 & 0.7426 & 0.8085 \\
Emory & 0.8974 & 0.8744 & 0.9216 & 0.9005 & Kai & 0.7772 & 0.7540 & 0.8019 & 0.7830 \\
NJU ConvAI & 0.8973 & 0.8751 & 0.9206 & 0.9005 & IBS & 0.7765 & 0.7692 & 0.7839 & 0.7870 \\
IDSOU & 0.8964 & 0.8988 & 0.8941 & 0.9025 & MrRobot & 0.7648 & 0.7515 & 0.7786 & 0.7740 \\
ComplexDataLab & 0.8945 & \textbf{0.9195} & 0.8708 & 0.9030 & ISWARA & 0.7631 & 0.8073 & 0.7235 & 0.7880 \\
UPennHLP & 0.8941 & 0.9028 & 0.8856 & 0.9010 & TheWalkingBy & 0.7614 & 0.7709 & 0.7521 & 0.7775 \\
DATAMAFIA & 0.8940 & 0.8857 & 0.9025 & 0.8990 & KZhu & 0.7580 & 0.7788 & 0.7383 & 0.7775 \\
NIT\_COVID-19 & 0.8914 & 0.8594 & 0.9258 & 0.8935 & IRLab@IITBHU & 0.7508 & 0.7904 & 0.7150 & 0.7760 \\
CXP949 & 0.8910 & 0.8698 & 0.9131 & 0.8945 & \underline{Baseline--fastText} & 0.7503 & 0.7730 & 0.7288 & 0.7710 \\
NHK\_STRL & 0.8898 & 0.8985 & 0.8814 & 0.8970 & Amrita\_CEN\_NLP & 0.7496 & 0.8078 & 0.6992 & 0.7795 \\
COVCOR20 & 0.8887 & 0.8655 & 0.9131 & 0.8920 & intelligentCyborgs & 0.7417 & 0.6507 & 0.8623 & 0.7165 \\
CIA\_NITT & 0.8887 & 0.8772 & 0.9004 & 0.8935 & BhagwanBharose & 0.7269 & 0.7723 & 0.6864 & 0.7565 \\
honeybee & 0.8884 & 0.8956 & 0.8814 & 0.8955 & IITKGPPHD & 0.7132 & 0.7535 & 0.6769 & 0.7430 \\
BANANA & 0.8881 & 0.8853 & 0.8909 & 0.8940 & NITK\_NLP & 0.6826 & 0.7581 & 0.6208 & 0.7275 \\
SU-NLP & 0.8881 & 0.8895 & 0.8867 & 0.8945 & 36H102 & 0.5800 & 0.5015 & 0.6875 & 0.5300 \\
VT & 0.8846 & 0.8723 & 0.8972 & 0.8895 & TMU-COVID19 & 0.5789 & 0.5000 & 0.6875 & 0.5280 \\
\hline
\end{tabular}
}
\caption{Final results on the test set. \textbf{P}, \textbf{R} and  \textbf{Acc.} denote the Precision, Recall and Accuracy, respectively. Teams are ranked by their highest  F$_1$ score.}
\label{tab:results}
\end{table*}

\section{Results}

In total, 121 teams spreading across 20 different countries registered to participate in our WNUT-2020 Task 2 during the system development phase. Of those 121 teams, 55 teams uploaded their submissions for the final evaluation phase.\footnote{CXP949 is not shown on our CodaLab leaderboard  because this team unfortunately makes an incorrectly-formatted submission file name, resulting in a fail for our CodaLab automatic evaluation program. We manually re-evaluate their submission and include its obtained results in Table \ref{tab:results}.}

We report results obtained for each team in Table \ref{tab:results}. The baseline fastText achieves 0.7503 in F$_1$ score. In particular, 48 teams outperform the baseline in terms of F$_1$. There are 39 teams with an F$_1$ greater than 0.80, in which  10 teams are with an F$_1$ greater than 0.90.  
Both  NutCracker \cite{nutcracker} and NLP\_North \cite{nlpnorth} obtain the highest F$_1$ score at 0.9096, in which NutCracker obtains the highest Accuracy at 91.50\% that is 0.1\% absolute higher than NLP\_North's. 

Of the 55 teams, 36 teams submitted their system paper, in which 34 teams' papers are finally included in the Proceedings. %\footnote{%Most of the system paper submissions are examined by 3 reviewers in which one of three reviewers is sampled from three senior members of the task organizer team, and the two remaining reviewers are senior members of other participating teams. 
%mazingAI withdraw their paper during the review process, and Loner's paper   could not meet the publication standards after two rounds of revision.} 
%\footnote{One participating team withdraw their paper during the review process, and another team's paper could not meet the publication standard after two rounds of revision.}  
% 
All of the 36 teams with paper submissions employ pre-trained language models to extract latent features for learning classifiers. The majority of  pre-trained language models employed include BERT \cite{devlin-etal-2019-bert}, XLNet \cite{NIPS2019_8812},  RoBERTa \cite{RoBERTa}, BERTweet \cite{bertweet} and especially CT-BERT \cite{muller2020covid}. 

Not surprisingly, CT-BERT, resulted in by continuing pre-training from the pre-trained BERT-large model on a corpus of 22.5M COVID-19 related Tweets, is utilized in a large number of the highly-ranked systems. In particular, all of top 6  teams including NutCracker, NLP\_North, UIT-HSE \cite{uit}, \#GCDH \cite{GCDH}, Loner and Phonemer \cite{Phonemer} utilize CT-BERT. That is why we find slight differences in their obtained F$_1$  scores. In addition, ensemble techniques are also used in a large proportion (61\%) of the participating teams. Specifically,  to obtain the best performance,  the top 10 teams, except NLP\_North, \#GCDH and Loner, all employ ensemble techniques.

\section{Conclusion}

In this paper, we have presented an overview of the WNUT-2020 Task 2 ``Identification of Informative COVID-19 English Tweets'': (i) Provide details of the task, data preparation process, and the task organization, and (ii) Report the results obtained by participating teams and outline their commonly adopted approaches.

We receive registrations from 121 teams and final system evaluation submissions from 55 teams, in which 34/55 teams contribute detailed system descriptions. The evaluation results show  that many  systems obtain a very high performance of up to 0.91 F$_1$ score on the task, using pre-trained  language models which are fine-tuned on unlabelled COVID-19 related Tweets (CT-BERT) and are subsequently trained on this task.

{%\footnotesize
\bibliographystyle{acl_natbib}
\bibliography{REFs}

\begin{thebibliography}{16}
\expandafter\ifx\csname natexlab\endcsname\relax\def\natexlab#1{#1}\fi

\bibitem[{Devlin et~al.(2019)Devlin, Chang, Lee, and
  Toutanova}]{devlin-etal-2019-bert}
Jacob Devlin, Ming-Wei Chang, Kenton Lee, and Kristina Toutanova. 2019.
\newblock {BERT}: Pre-training of deep bidirectional transformers for language
  understanding.
\newblock In \emph{Proceedings of the 2019 Conference of the North American
  Chapter of the Association for Computational Linguistics: Human Language
  Technologies, Volume 1 (Long and Short Papers)}, pages 4171--4186.

\bibitem[{Fleiss(1971)}]{fleiss1971measuring}
Joseph~L Fleiss. 1971.
\newblock Measuring nominal scale agreement among many raters.
\newblock \emph{Psychological bulletin}, 76(5):378--382.

\bibitem[{Joulin et~al.(2017)Joulin, Grave, Bojanowski, and
  Mikolov}]{joulin2017bag}
Armand Joulin, Edouard Grave, Piotr Bojanowski, and Tomas Mikolov. 2017.
\newblock {Bag of Tricks for Efficient Text Classification}.
\newblock In \emph{Proceedings of the 15th Conference of the European Chapter
  of the Association for Computational Linguistics: Volume 2, Short Papers},
  pages 427--431.

\bibitem[{Kumar and Singh(2020)}]{nutcracker}
Priyanshu Kumar and Aadarsh Singh. 2020.
\newblock {NutCracker at WNUT-2020 Task 2: Robustly Identifying Informative
  COVID-19 Tweets using Ensembling and Adversarial Training }.
\newblock In \emph{Proceedings of the 6th Workshop on Noisy User-generated
  Text}.

\bibitem[{Lamsal(2020)}]{datacovid}
Rabindra Lamsal. 2020.
\newblock \href {https://doi.org/http://dx.doi.org/10.21227/781w-ef42}
  {{CORONAVIRUS (COVID-19) TWEETS DATASET}}.
\newblock \emph{IEEE Dataport}.

\bibitem[{Landis and Koch(1977)}]{landis1977measurement}
J~Richard Landis and Gary~G Koch. 1977.
\newblock The measurement of observer agreement for categorical data.
\newblock \emph{Biometrics}, 33(1):159--174.

\bibitem[{Liu et~al.(2019)Liu, Ott, Goyal, Du, Joshi, Chen, Levy, Lewis,
  Zettlemoyer, and Stoyanov}]{RoBERTa}
Yinhan Liu, Myle Ott, Naman Goyal, Jingfei Du, Mandar Joshi, Danqi Chen, Omer
  Levy, Mike Lewis, Luke Zettlemoyer, and Veselin Stoyanov. 2019.
\newblock {RoBERTa: {A} Robustly Optimized {BERT} Pretraining Approach}.
\newblock \emph{arXiv preprint}, arXiv:1907.11692.

\bibitem[{M{\o}ller et~al.(2020)M{\o}ller, van~der Goot, and Plank}]{nlpnorth}
Anders~Giovanni M{\o}ller, Rob van~der Goot, and Barbara Plank. 2020.
\newblock {NLP North at WNUT-2020 Task 2: Pre-training versus Ensembling for
  Detection of Informative COVID-19 English Tweets}.
\newblock In \emph{Proceedings of the 6th Workshop on Noisy User-generated
  Text}.

\bibitem[{M{\"u}ller et~al.(2020)M{\"u}ller, Salath{\'e}, and
  Kummervold}]{muller2020covid}
Martin M{\"u}ller, Marcel Salath{\'e}, and Per~E Kummervold. 2020.
\newblock {COVID-Twitter-BERT: A Natural Language Processing Model to Analyse
  COVID-19 Content on Twitter}.
\newblock \emph{arXiv preprint arXiv:2005.07503}.

\bibitem[{Nakayama et~al.(2018)Nakayama, Kubo, Kamura, Taniguchi, and
  Liang}]{doccano}
Hiroki Nakayama, Takahiro Kubo, Junya Kamura, Yasufumi Taniguchi, and Xu~Liang.
  2018.
\newblock \href {https://github.com/doccano/doccano} {{{doccano}: Text
  Annotation Tool for Human}}.
\newblock Software available from https://github.com/doccano/doccano.

\bibitem[{Nguyen et~al.(2020)Nguyen, Vu, and Nguyen}]{bertweet}
Dat~Quoc Nguyen, Thanh Vu, and Anh~Tuan Nguyen. 2020.
\newblock {BERTweet: A pre-trained language model for English Tweets}.
\newblock In \emph{Proceedings of the 2020 Conference on Empirical Methods in
  Natural Language Processing: System Demonstrations}.

\bibitem[{Tran et~al.(2020)Tran, Phan, Nguyen, and Nguyen}]{uit}
Khiem Tran, Hao Phan, Kiet Nguyen, and Ngan Luu~Thuy Nguyen. 2020.
\newblock {UIT-HSE at WNUT-2020 Task 2: Exploiting CT-BERT for Identifying
  COVID-19 Information on the Twitter Social Network}.
\newblock In \emph{Proceedings of the 6th Workshop on Noisy User-generated
  Text}.

\bibitem[{Varachkina et~al.(2020)Varachkina, Ziehe, Do\'nicke, and
  Pannach}]{GCDH}
Hanna Varachkina, Stefan Ziehe, Tillmann Do\'nicke, and Franziska Pannach.
  2020.
\newblock {\#GCDH at WNUT-2020 Task 2: BERT-Based Models for the Detection of
  Informativeness in English COVID-19 Related Tweets}.
\newblock In \emph{Proceedings of the 6th Workshop on Noisy User-generated
  Text}.

\bibitem[{Wadhawan(2020)}]{Phonemer}
Anshul Wadhawan. 2020.
\newblock {Phonemer at WNUT-2020 Task 2: Sequence Classification Using COVID
  Twitter BERT and Bagging Ensemble Technique based on Plurality Voting}.
\newblock In \emph{Proceedings of the 6th Workshop on Noisy User-generated
  Text}.

\bibitem[{{Wang} et~al.(2011){Wang}, {Li}, and {Fe}}]{5767865}
J.~{Wang}, G.~{Li}, and J.~{Fe}. 2011.
\newblock Fast-join: An efficient method for fuzzy token matching based string
  similarity join.
\newblock In \emph{Proceedings of the 27th IEEE International Conference on
  Data Engineering}, pages 458--469.

\bibitem[{Yang et~al.(2019)Yang, Dai, Yang, Carbonell, Salakhutdinov, and
  Le}]{NIPS2019_8812}
Zhilin Yang, Zihang Dai, Yiming Yang, Jaime Carbonell, Russ~R Salakhutdinov,
  and Quoc~V Le. 2019.
\newblock {XLNet: Generalized Autoregressive Pretraining for Language
  Understanding}.
\newblock In \emph{Advances in Neural Information Processing Systems 32}, pages
  5753--5763.

\end{thebibliography}
}

\end{document}